%% file: main.tex
\documentclass[letterpaper]{article}
\usepackage{aaai20}
\usepackage{times}
\usepackage{helvet}
\usepackage{courier}
\usepackage[hyphens]{url}
\usepackage{graphicx}
\usepackage{graphicx}
\usepackage[vlined,ruled,linesnumbered]{algorithm2e}
\usepackage{array}
\usepackage{multirow}

\title{Adversarial Model Extraction on Graph Neural Networks}

\author{David DeFazio and Arti Ramesh\\
Binghamton University\\
ddefazi1@binghamton.edu,
artir@binghamton.edu
}

\begin{document}
\maketitle
\begin{abstract}

Along with the advent of deep neural networks came various methods of exploitation, such as fooling the classifier or contaminating its training data. Another such attack is known as model extraction, where provided API access to some black box neural network, the adversary extracts the underlying model. This is done by querying the model in such a way that the underlying neural network provides enough information to the adversary to be reconstructed. While several works have achieved impressive results with neural network extraction in the propositional domain, this problem has not yet been considered over the relational domain, where data samples are no longer considered to be independent and identically distributed (iid). Graph Neural Networks (GNNs) are a popular deep learning framework to perform machine learning tasks over relational data. In this work, we formalize an instance of GNN extraction, present a solution with preliminary results, and discuss our assumptions and future directions.

\end{abstract}

\input{introduction}
\input{related_work}
\input{gnn_model_extraction}

\input{discussion}
\input{experiments}

\input{future_work}

\bibliographystyle{aaai}
\bibliography{references}
\end{document}

%% file: introduction.tex
\section{Introduction}

In recent years, there has been substantial effort in applying deep learning models to graphs, culminating in a variety of different GNN architectures \cite{kipf2017semi,hamilton2017inductive,velickovic2018graph,xu2018how}. With these new models, come new adversarial approaches to decrease performance in production models (evasion attack) or contaminate training data in an undetectable way (poisoning attack). In this study, we attempt a different form of attack, known as \textit{model extraction}. This is where an adversary tries to steal an underlying model, given only API access to the victim model's predicted labels. 

In a relational setting, we additionally need to address issues associated with querying the model. Unlike neural network extraction on iid data, we need to modify the underlying graph in order to obtain new label information from the victim. Graph modifications can be in the form of  adding or deleting adversarial nodes, as well as modifying existing node's features. This is typically more expensive for the model provider, more effort for the adversary, and more suspicious when repeated a large number of times when compared to the propositional setting.

For example, consider a GNN trained for the task of node classification (predicting a label for each node) on a social network. Typically, individuals are able to modify their own node's features and edges, and create a limited number of new nodes. Without the ability to query the model a significantly large number of times, extraction becomes more difficult. Additionally, we must consider typical properties of social networks. One common property is homophily: users naturally fall into different communities. This further increases the difficulty of model extraction, as GNNs build a node's latent representation according to the features of its neighbors. Adding a handful of adversarial nodes with features representative of a different class might not significantly alter the original neighborhood's feature distribution, causing the underlying victim GNN to return the same labels for this neighborhood, and thus be uninformative for the task of model extraction.

In this paper, we follow a learning based extraction approach, which we extend to GNNs. This strategy involves learning a new model which trains on data labeled by the victim. More specifically, we learn our model extraction by iteratively perturbing a subgraph of the victim model's original graph, and training our extraction over different instances of these perturbed subgraphs, whose nodes are labeled by the victim GCN. We make the following contributions:

\begin{enumerate}

	\item We introduce the novel problem of GNN model extraction, present a preliminary approach for model extraction, analyze various considerations and assumptions in the approach, and discuss directions for future progress.

	\item We learn graph convolutional network (GCN) extractions of victims trained on two datasets: Cora and Pubmed. Our extraction achieves approximately 80\% fidelity despite having access to only a subgraph of the entire training graph.

\end{enumerate}

Our empirical evaluation is currently limited, as extending existing approaches that extract non relational models to our setting is not trivial. We could consider a query to our black box GNN as adding an edgeless node, which would allow for independent queries. This is highly suspicous however, as this requires adding too many edgeless nodes, which is not typical behavior. Therefore, we cannot find any suitable baselines for comparison, and exclude baselines from our evaluation.

The remainder of this paper focuses on our preliminary algorithm and results for learning a high fidelity GNN extraction, along with an extended discussion on our assumptions, limitations of our approach, and future directions.

%% file: related_work.tex

\section{Related Work}

Model extraction has been studied in a variety of different contexts. One of the earlier works by \citeauthor{tramer2016stealing} \shortcite{tramer2016stealing} defines extraction attacks such as equation solving on logistic regression, and path finding on decision trees. \citeauthor{orekondy2019knockoff} \shortcite{orekondy2019knockoff} develop a reinforcement learning approach to select samples to train the extraction on. \citeauthor{oh2019towards} \shortcite{oh2019towards} learn a meta-learner over different neural network architectures to extract the victim's architecture. \citeauthor{wang2018stealing} \shortcite{wang2018stealing} develop an approach to steal the hyperparameters of a model when given access to the training dataset and ML algorithm. \citeauthor{milli2018model} \shortcite{milli2018model} develop a method of model extraction using gradient explanations. Less victim queries are needed when provided input gradients, as they are more informative than labels. \citeauthor{jagielski2019high} \shortcite{jagielski2019high} discuss limitations of learning based approaches, and introduces a functionally-equivalent extraction attack on a neural network with 1 hidden layer. None of these approaches considers non-iid data.

Another interesting work by \citeauthor{chandrasekaran2020exploring} \shortcite{chandrasekaran2020exploring} shows that the model extraction setting is equivalent to active learning in the query synthesis setting, where the learner can generate any instance within the input space and request a label from an oracle. This means that solutions from active learning can be (and have been) applied to the task of model extraction. Again, these active learning approaches are designed for non-relational data. While some works do study active learning for GNNs \cite{wu2019active,cai2017active,gao2018active}, we leave active learning solutions for GNN extraction as future work.

There has been progress in adversarial attacks on GNNs. \citeauthor{zugner2018adversarial} \shortcite{zugner2018adversarial} perform a poisoning attack, where they are cautious of the perturbing degree and feature distributions to avoid detection. This is relevant to our problem, where we want to retrieve different subgraph samples without being detected as malicious.

%% file: gnn_model_extraction.tex
\section{Model Extraction of GNNs}

\subsection{Problem Statement}

We aim to solve the following: Given API access to some GNN performing node classification on graph $G = (A,X)$, learn a high fidelity extraction of this GNN. 

More specifically, we have access to a small subgraph $G_s = (A_{s}, X_{s}) \subset G$, where $A_s$ is the adjacency matrix and $X_s$ contains a feature vector of each node in its rows of our subset. We have the ability to change features, edges, and add new adversarial nodes in $G_s$, but can't access other nodes in $G$. We can use the victim model $f$ to get the labels of nodes in $G_s$. We want to perturb $G_s$ a minimal number of times, in an undetectable manner, such that we learn enough about $f$ to learn a high fidelity extraction $g$.

\subsection{Subgraph Sampling Algorithm}

Algorithm \ref{alg:extraction} provides pseudocode for learning our extracted model $g$. We assume multinomial distributions $F$ and $M$ as prior knowledge, and consider the case where each node has sparse binary features, as is true in citation networks. We first make samples of $G_s$ by sampling its number and type of features for each node in $G_s$ from our prior distributions for each class, as defined in \textit{GenerateSample()}. We do this to capture global information of the graph. For a high fidelity extraction, we need to train our extraction on a graph which contains labels from each class predicted by the victim. We also add edges to our subgraph such that it is complete. This is motivated in our discussion. We next save each node's new labels from $f$, along with the 2-hop adjacency matrix and feature matrix from our sample of $G_s$. We train $g$ over our set of sampled adjacency matricies, by combining the sampled adjacency matricies into one block diagonal matrix for training over multiple graphs \cite{kipf2017semi}. We use cross-entropy loss to train our extracted model, which we discuss in further detail in the next section.

\subsection{Approximating Inaccessible Neighbors}

An issue with Algorithm \ref{alg:extraction} is that our extraction $g$ is trained on labels that are based on information not present in our graph samples. This is because victim labels of nodes not in the center of our sample are also predicted by information from inaccessible nodes (3 and 4 hops away from the center of our 2-hop subgraph). We extend our algorithm to approximate these inaccessible nodes for each subgraph we generate. This approximation comes at no additional cost in terms of stealth or victim queries, as we are only adding these nodes to our internal training samples, not the original graph. We provide pseudocode in Algorithm \ref{alg:node_approx}. Note that we use $\rho=2$ in our experiments, and each new node is connected to multiple nodes in our sample.

%% file: discussion.tex

\subsection{Analyzing Modeling Considerations}

In this section, we discuss various aspects of our algorithm and the potential limitations of our assumptions. We also note some interesting empirical observations that guide our approach.

\subsubsection{Variance of Feature Distributions}

Our graph samples consist of nodes sampled from the same multinomial distribution. This assumes that each node of a class can be characterized by a single multinomial distribution over features, which likely is not the case in real-world scenarios. Because these feature distributions are constructed by prior knowledge, there is likely a bias towards certain groups of features.

To introduce noise, for each feature we sample from $F_c$, there is a probability $\epsilon$ that we select a different random feature. While our preliminary results indicate that fidelity is highest with no noise ($\epsilon = 0$), we suspect introducing noise could help in better capturing information not encoded in our prior distributions.

 \begin{algorithm}

\DontPrintSemicolon

\KwIn{2-hop subgraph $G_s$\; 
API access to victim $f$\;
Each class's feature distribution $F$\;
Each class's number of features distribution $M$\;
Number of graph samples per class $n$\;
Number of classes $C$
}

\KwOut{Extracted model $g$}

$S_{A} = \{A_{s}\}$\;
$S_{X} = \{X_{s}\}$\;
$S_{L} = \{f(G_{s})\}$\;

\For{$i \gets 1$ \textbf{to} $n$}{

	\For{$c \gets 1$ \textbf{to} $C$}{

		$A_{c}, X_{c}$ = $GenerateSample$($F_{c}$, $M_{c}$, $A_{s}$)\;
		 $S_{A} = S_{A} \cup A_{c}$\;
		 $S_{X} = S_{X} \cup X_{c}$\;
		 Perturb nodes in $G$ according to $A_{c}$ and $X_{c}$\;
		 $S_{L} = S_{L} \cup f(G_{s})$\;

	}

}
$A_{G}$ = diagonal block matrix with all elements in $S_{A}$\;
$X_{G}$ = row-wise concatenation of all elements in $S_{X}$\;
Train 2-layer GCN $g$, given adjacency $A_{G}$, features $X_{G}$, and labels $S_{L}$\;

	 \Return{$g$}
	 \;
	 \;
\SetKwFunction{FMain}{GenerateSample}
\SetKwProg{Pn}{Function}{:}{\KwRet}
  \Pn{\FMain{$F_{c}$, $M_{c}$, $A_{s}$}}{
  
	$A_{c}$ = Complete graph with same nodes as$A_{s}$\;
	$X_{c} = \mathbf{0}$\;

		\For{$i \gets 1$ \textbf{to} $|A_{s}|$}{

			$m \sim M_c$\;

			\For{$j \gets 1$ \textbf{to} $m$}{
	
				$x \sim F_c$\;
				$X_{c}[i,x] = 1$\;
	
			}

		}

	\Return{$A_{c},X_{c}$}
  }

 \caption{Algorithm for extracting model $g$, given a 2-hop subgraph, and API access to $f$}
 \label{alg:extraction}
\end{algorithm}

 \begin{algorithm}

\DontPrintSemicolon

\SetKwFunction{FMain}{ApproximateNodes}
\SetKwProg{Pn}{Function}{:}{\KwRet}
  \Pn{\FMain{$G_{c}$}}{
  
  	$L$ = $f(G_{c})$\;
	$center$ = center node in $G_c$\;
	
	\For{$n \in G_{c} \setminus center$}{
		$D_n$ = $L_{n} - L_{center}$\; 
	}
	
	\For{$c \in C$}{
	
		\uIf{$D_{nc} > 0$}{
			Each subgraph node needs $D_{nc} * \rho * Degree(n)$ nodes added per class.
  		}
		
	}

	Create new nodes for each non-center subgraph node.

	\Return{$G_{c}$}
  }

 \caption{Approximate inaccessible nodes for samples that extraction $g$ will train on.}
 \label{alg:node_approx}
\end{algorithm}

\subsubsection{Loss}

In this work, we assume that the victim model returns a probability distribution over classes. Our extracted model's loss function should then use the victim's probability distributions as the true label. We use cross-entropy loss:

$$ \mathcal{L} = -\sum_{ n \in G_s} \sum_{c=1}^{C} f(n)_{c} \ln g(n)_{c}$$

where $f(n)_{c}$ is the victim's probability of node $n$ belonging to class $c$. Surprisingly, our preliminary results indicate that our extracted model has higher fidelity when we consider $f(n)$ as a one-hot vector, with the largest probability class as the prediction. This is unintuitive, as we expect runner-up class probabilities to provide valuable insight as to how the victim makes predictions.

We also consider the case where we know the true labels of the dataset, and balance training the model between extracting a high fidelity model, and an accurate model:

$$ \mathcal{L} =  - \sum_{ n \in G_s} \sum_{c=1}^{C} Y_{nc} \ln g(n)_{c} - \sum_{ n \in G_s} \sum_{c=1}^{C} f(n)_{c} \ln g(n)_{c} $$

where $Y_{nc} \in \{0,1\}$ is $1$ if node $n$ belongs to class $c$. Being that the victim model is trained on true labels, our intuition is that we want our extracted model to follow a similar procedure, in the hopes of learning a similar model. Our results so far indicate that including true labels does not improve fidelity.



\subsection{Analyzing Assumptions}

Below we discuss our assumptions, their limitations, and potential to relax them.

\subsubsection{Stealth}

We make the assumption that we can perturb our 2-hop graph in any way, going undetected. Our results indicate that for a simple 2 layer GCN extraction on Cora, for some nodes, it takes up to 100 samples per class, or 700 total samples just to achieve 82\% fidelity. Aside from suspicion raised due to the number of perturbations, each instance of our graph perturbations are highly suspicious as well. Perturbing the graph for one sample involves severely altering the degree distribution and feature distributions of nodes within our 2-hop subgraph. In particular, line 17 of Algorithm \ref{alg:extraction}  adds edges to our 2-hop window until it is a complete subgraph. We make such perturbations in order to outweigh the influence of nodes 3 and 4 hops away, which we have no access to. Otherwise, if our subgraph is sparsely connected, the victim's aggregation function might receive greater influence from inaccessible nodes, and thus their labels would not capture enough global graph information that our perturbations are attempting to represent.

 \citeauthor{zugner2018adversarial} \shortcite{zugner2018adversarial} perform a poisoning attack with less noticeable perturbations. For example, when perturbing the graph, they ensure that the degree distribution still closely follows the power law distribution (a typical degree distribution that graphs tend to naturally follow). Additionally, they set a limit on the number of allowed perturbations, and preserve feature co-occurrence. While we are operating with access to a significantly limited portion of the graph, these ideas can still be extended to GNN extraction attacks. 

\subsubsection{Architecture}

Our assumption of a 2-layer GCN is overly simplifying and this might not be the case in production. While \citeauthor{oh2019towards} \shortcite{oh2019towards} learns the model architecture for neural networks on iid data with a meta-learner, extending this approach to GNNs is not trivial, due to issues we face with querying our relational victim, compared to a victim operating on iid data.

\subsubsection{Access to Node Neighbors}

Our assumption of having full access to a 2-hop subgraph can be true in theory, but would be unrealistic in practice. It is common for people to have hundreds of friends in a social network. Therefore, we would need $\approx$ 10,000 cooperating users just for a 2-hop subgraph. To circumvent this, one idea we are considering in the near future is to learn our extraction entirely based on a disjoint subgraph of adversarial nodes we add. This has many additional implications, which we leave for future work.

\subsubsection{Priors}

Prior knowledge of each class's feature distribution is another simplifying assumption. Consider the case where we have a large citation network with 1000 classes. In this case it might be too unrealistic or expensive to obtain good prior feature distributions for each class.
 

%% file: experiments.tex
\section{Experiments}

We train our victim GCN $f$ on the Cora and Pubmed citation network datasets \cite{sen2008collective}, for the task of node classification: 

\begin{enumerate}

	\item \textbf{Cora} contains 2708 labeled nodes, each of which has 1 of 7 possible labels. We train our victim on 140 of these nodes.
	
	\item \textbf{Pubmed} contains 19717 labeled nodes, each of which has 1 of 3 possible labels. We train our victim on 60 of these nodes.

\end{enumerate}

Each node represents a scientific paper, and each binary feature determines whether some dictionary word is in the paper or not. Two papers are linked if one of the papers cites the other. Many GNN architectures consider the graph as undirected by default. We therefore consider our citation networks as undirected, as a convienience. Although our algorithm works with any GNN architecture, in this work we focus on GCN. Note that we can't compare against baselines, as GNN extraction has not previously been considered.

Next, we randomly select a node with a 2-hop neighborhood size of at least 10, but no larger than 150. A subgraph with fewer than 10 nodes does not contain enough information, while a subgraph larger than 150 nodes takes longer to extract a model from, and is more suspicious. We get our multinomial prior distributions $F$ and $M$ by taking the frequency of each feature per class, and number of features per node and class respectively, and normalize. We assume access to this node's 2-hop subgraph, and we run our extraction algorithm to train $g$.

Once we train $g$ on our different samples of $G_s$, we measure its fidelity to $f$ by predicting over the entire original graph $G$, and determining what percentage of labels $g$ predicted the same as $f$. Note that this is a different task from learning an accurate model. We learn $g$ strictly to match the predictions of $f$, regardless of whether $f$ is making accurate predictions.
\begin{table}[!ht]
\begin{tabular}{ | m{2cm} |m{2cm}|m{2cm}| } 
 \hline
 Node ID & Number Subgraph Nodes & Fidelity \\ 
 \hline
 503 & 17.0 & 0.75 \\ 
 \hline
 1174 & 32.0 & 0.80 \\ 
 \hline
 102 & 23.0 & 0.79 \\ 
 \hline
 738 & 116.0 & 0.80 \\ 
 \hline
 1977 & 21.0 &  0.80 \\ 
 \hline
 2218 & 11.0 &  0.64 \\ 
 \hline
 321 & 141.0 &  0.78 \\ 
 \hline
 560 & 59.0 & 0.80 \\ 
 \hline
 343 & 23.0 & 0.79 \\ 
 \hline
 2317 & 16.0 & 0.75 \\ 
 \hline
 AVG & 45.9 & 0.77 \\ 
 \hline 
\end{tabular}

  \caption{Fidelity of extracted model $g$, when varying the accessible subgraph. Using 10 samples per class on Cora dataset.}
 \label{tabel:cora_nodes} 
\end{table}






\begin{table}[ht!]

\begin{tabular}{ |m{1cm}|m{0.8cm}|m{2cm}|m{1.3cm}|m{1cm}| } 
 \hline
 Dataset & Node ID & Number Subgraph Nodes & Samples per Class & Fidelity \\ 
 \hline
 \multirow{10}{*}{Cora} & 675 & 48 & 1 & 0.67 \\ \cline{2-5}
  & 675 & 48 & 3 & 0.79 \\ \cline{2-5}
  & 675 & 48 & 5 & 0.82 \\ \cline{2-5}
  & 675 & 48 & 10 & 0.81 \\ \cline{2-5}
  & 675 & 48 & 100 & 0.83 \\ \cline{2-5}
  & 1956 & 13 & 1 & 0.36 \\ \cline{2-5}
  & 1956 & 13 & 3 & 0.71 \\ \cline{2-5}
  & 1956 & 13 & 5 & 0.75 \\ \cline{2-5}
  & 1956 & 13 & 10 & 0.81 \\ \cline{2-5}
  & 1956 & 13 & 100 & 0.82 \\
  \hline
 \multirow{10}{*}{Pubmed} & 3028 & 16 & 1 & 0.67 \\ \cline{2-5}
  & 3028 & 16 & 7 & 0.72 \\ \cline{2-5}
  & 3028 & 16 & 11 & 0.76 \\ \cline{2-5}
  & 3028 & 16 & 23 & 0.80 \\ \cline{2-5}
  & 3028 & 16 & 233 & 0.83 \\ \cline{2-5}
  & 12759 & 75 & 1 & 0.72 \\ \cline{2-5}
  & 12759 & 75 & 7 & 0.80 \\ \cline{2-5}
  & 12759 & 75 & 11 & 0.80 \\ \cline{2-5}
  & 12759 & 75 & 23 & 0.81 \\ \cline{2-5}
  & 12759 & 75 & 233 & 0.82 \\
 \hline
 
\end{tabular}

  \caption{Results on varying the samples per class over the same nodes. Note that Pubmed has only 3 classes, which means we need more Pubmed samples per class to match the number of samples our algorithm is using to train $g$ on Cora.}
 \label{tabel:cora_pubmed_vary_samples}

\end{table}

Table \ref{tabel:cora_nodes} contains results over 10 random nodes in Cora, where we use 10 samples per class, or 70 total samples. A sample refers to a particular perturbation of the 2-hop graph, which represents some class, and is present in the block diagonal matrix. We must have at least 1 sample per class in order for our extraction to be able to learn over all classes. This table indicates that access to small portions of the graph can achieve up to 80\% fidelity, when following our strong assumptions. This helps us understand that a small number of subgraph samples each representing a different class is enough to possibly perform a successful model extraction. Our average subgraph size of 45.9 nodes consists of 1.7\% of all nodes in Cora.

Table \ref{tabel:cora_pubmed_vary_samples} varies the number of samples per class, keeping the node constant. Our results confirm that our extraction achieves higher fidelity with more samples. Additionally, we see that for larger subgraphs we have a relatively high fidelity for few samples, which only gradually improves with more samples. The reverse is true for nodes with small subgraphs; we have a low fidelity with few samples, which dramatically improves with more samples. Experiments with other nodes confirm this pattern. From these results, we see that when the victim provides more label information, whether its through larger subgraphs, or more samples, fidelity improves up to $\approx$ 80\% fidelity.

%% file: future_work.tex

\section{Conclusion and Future Work}

We present a preliminary case study of model extraction in GNNs. We present a simple solution that can achieve approximately 80\% fidelity on 2 citation networks. We also provide a detailed discussion of different assumptions and modeling considerations. Weakening these assumptions, and applying these ideas can potentially improve our attack, such that it is more realistic in practice, and achieves a higher fidelity extraction. Below, we present new directions to consider:


\begin{enumerate}

	\item \citeauthor{jagielski2019high} \shortcite{jagielski2019high} discuss a functionally equivalent approach to model extraction, which avoids the fundamental issue of non-determinisim plauged by learning based approaches. We plan to study how this can be extended to GNNs.

	\item We are training our GCN on a fake graph of samples which is substantially different from what we are testing on to measure fidelity. GCN is transductive, and thus has not been designed for the task of predicting unseen nodes well. This is motivation to experiment with inductive GNNs more suitable to predicting unseen nodes, such as GraphSAGE \cite{hamilton2017inductive}.
	
	\item We need a scalable model extraction approach, that uses a sufficiently small portion of the graph with a limited number of perturbations. Although our experiments use a small fraction of nodes in the dataset (up to 0.38\% in Pubmed), this percentage increases with denser graphs. Furthermore, obtaining accurate prior distributions over features is more costly as the number of nodes and classes increases.
	
	\item Consider the case where our GNN provides explanations along with our predictions, using a technique such as GNN Explainer \cite{ying2019gnn}. Just as \citeauthor{milli2018model} \shortcite{milli2018model} discuss how to take advantage of gradient explanations for model extraction attacks, we can consider how to use GNN explanations to improve fidelity.
	
	\item In this work, we only consider stealing GNNs for the task of node classification. We plan to extend this to other tasks such as link prediction and graph classification.

\end{enumerate}